\documentclass[runningheads]{llncs}
\usepackage{url}
\usepackage{graphicx}
\usepackage{xcolor}
\usepackage{subfig}
\begin{document}
\title{Automatic alignment of surgical videos \\ using kinematic data}
%
%
\author{Hassan Ismail Fawaz\textsuperscript{1}, 
Germain Forestier\textsuperscript{1,2}, 
Jonathan Weber\textsuperscript{1}, \\
Fran\c{c}ois Petitjean\textsuperscript{2},
Lhassane Idoumghar\textsuperscript{1} \and
Pierre-Alain Muller\textsuperscript{1}
}
\authorrunning{H. Ismail Fawaz et al.}
%
\institute{\textsuperscript{1}IRIMAS, University of Haute-Alsace, Mulhouse, France\\
\textsuperscript{2}Faculty of Information Technology, Monash University, Melbourne, Australia}
%
\maketitle              
\begin{abstract}
Over the past one hundred years, the classic teaching methodology of ``see one, do one, teach one'' has governed the surgical education systems worldwide.
With the advent of Operation Room 2.0, recording video, kinematic and many other types of data during the surgery became an easy task, thus allowing artificial intelligence systems to be deployed and used in surgical and medical practice.
Recently, surgical videos has been shown to provide a structure for peer coaching enabling novice trainees to learn from experienced surgeons by replaying those videos.
However, the high inter-operator variability in surgical gesture duration and execution renders learning from comparing novice to expert surgical videos a very difficult task.
In this paper, we propose a novel technique to align multiple videos based on the alignment of their corresponding kinematic multivariate time series data. 
By leveraging the Dynamic Time Warping measure, our algorithm synchronizes a set of videos in order to show the same gesture being performed at different speed.
We believe that the proposed approach is a valuable addition to the existing learning tools for surgery.

\end{abstract}

\section{Introduction}


\bigskip

\begin{figure}[t]
    \centering
    \subfloat[Video without alignment]{
    \includegraphics[width=.45\linewidth]{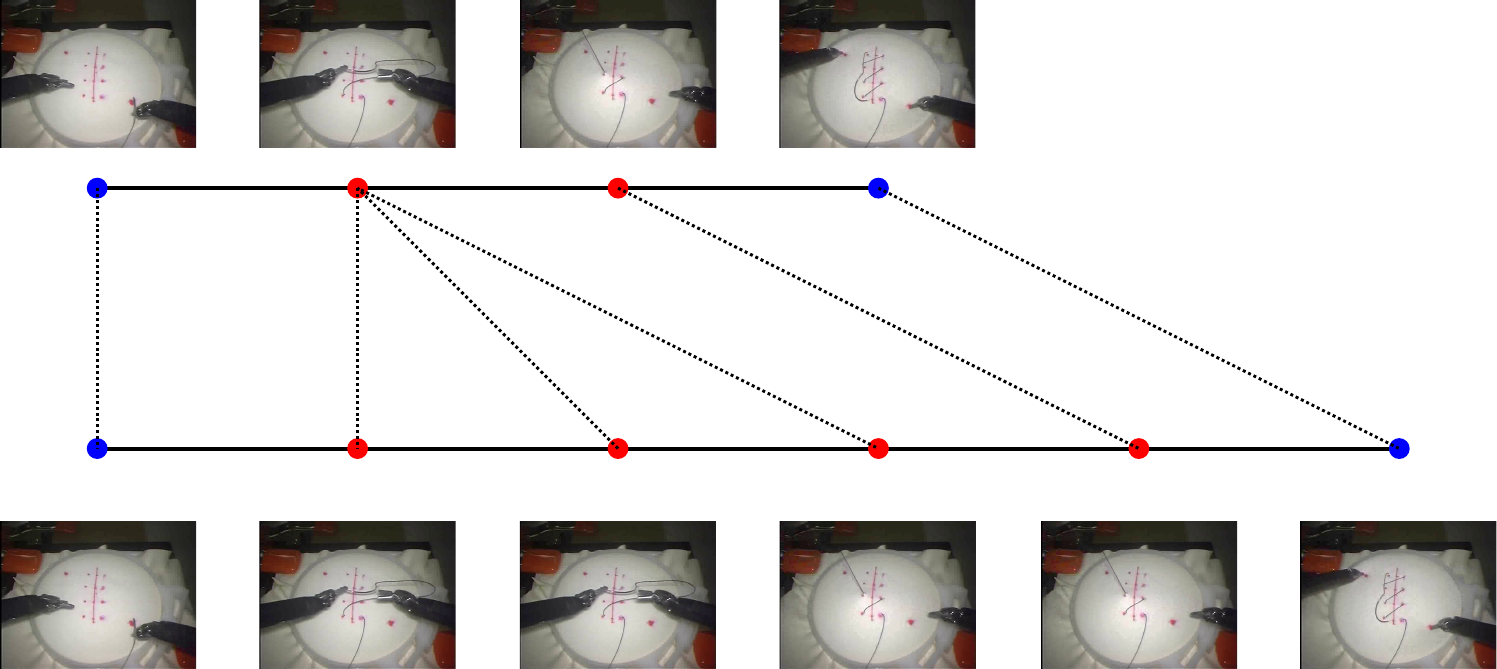}
    \label{sub:vid}
    }
    \subfloat[Video with alignment]{
    \includegraphics[width=.45\linewidth]{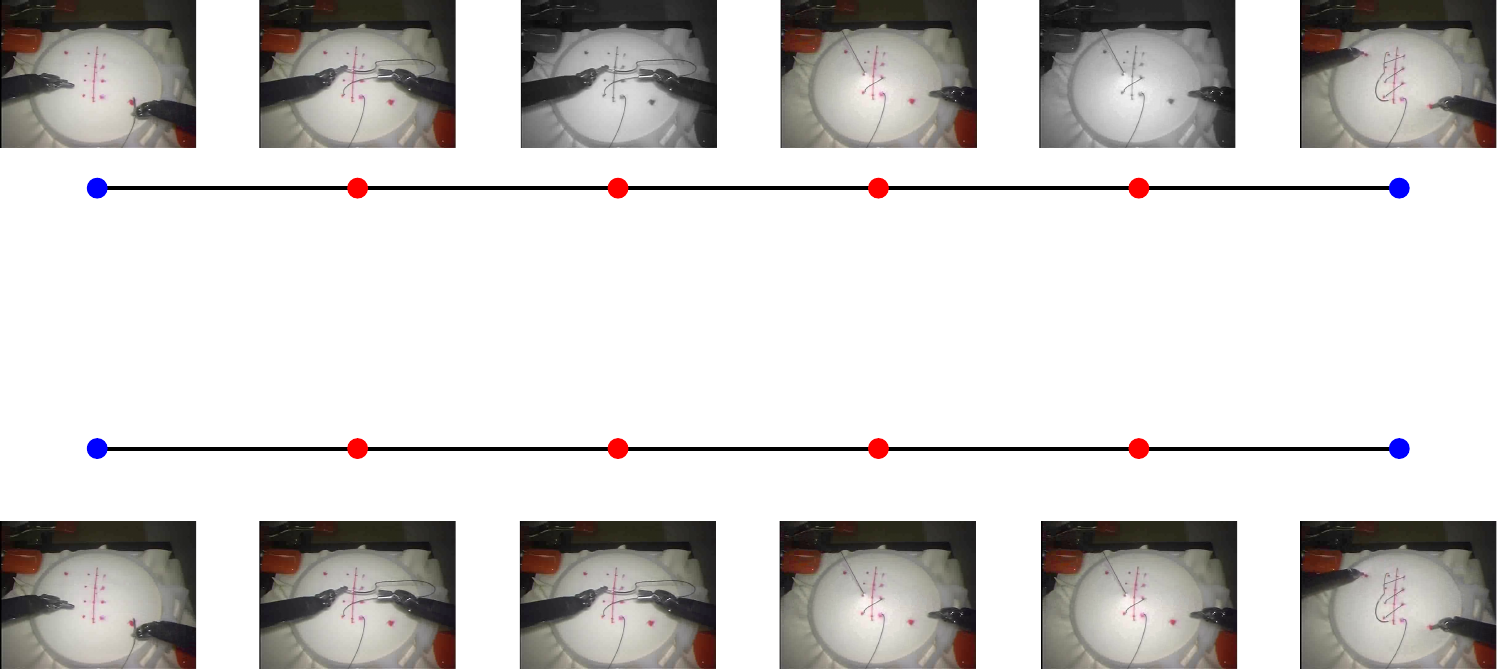}
      \label{sub:vid-synch}
      }
    \caption{Example on how a time series alignment is used to synchronize the videos by duplicating the gray-scale frames. Best viewed in color.}
    \label{fig:vid-synch}
\end{figure}

\begin{figure}[t]
    \centering
    \subfloat[Original time series without alignment]{
    \includegraphics[width=.65\linewidth]{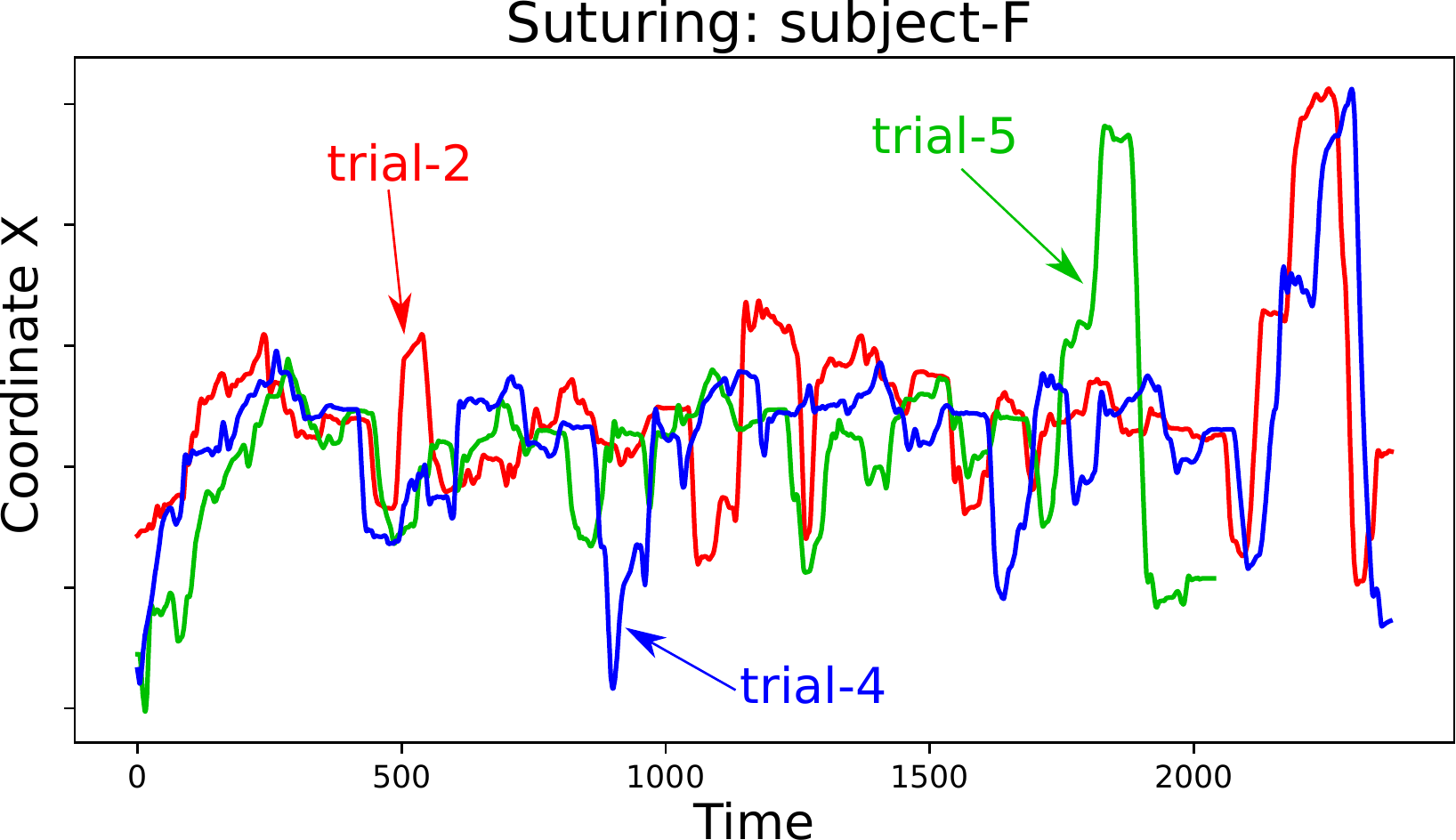}
    }\\
    \subfloat[Warped time series with alignment]{
 \includegraphics[width=.65\linewidth]{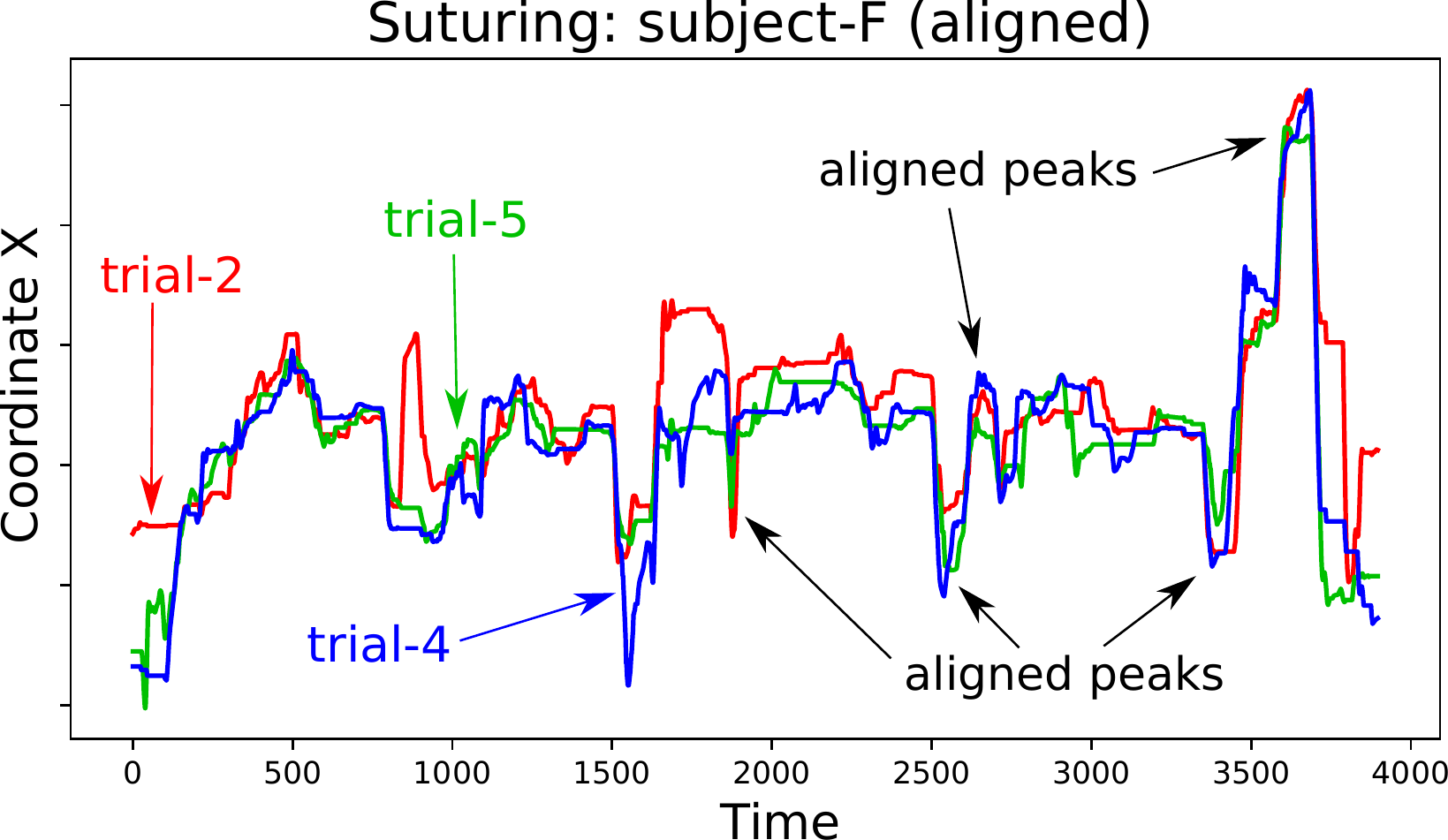}
      }
    \caption{Example of aligning coordinate X's time series for subject F, when performing three trials of the suturing surgical task.}
    \label{fig:trials}
\end{figure}

Educators have always searched for innovative ways of improving apprentices' learning rate.
While classical lectures are still most commonly used, multimedia resources are becoming more and more adopted~\cite{smith2000digital} especially in Massive Open Online Courses (MOOC)~\cite{means2009evaluation}.
In this context, videos have been considered as especially interesting as they can combine images, text, graphics, audio and animation.
The medical field is no exception, and the use of video-based resources is intensively adopted in medical curriculum~\cite{masic2008learning} especially in the context of surgical training~\cite{kneebone2002innovative}.
The advent of robotic surgery also simulates this trend as surgical robots, like the Da Vinci~\cite{davinci}, generally record video feeds during the intervention.
Consequently, a large amount of video data has been recorded in the last ten years~\cite{rapp2016youtube}.
This new source of data represent an unprecedented opportunity for young surgeons to improve their knowledge and skills~\cite{gao2014jhu}.
Furthermore, video can also be a tool for senior surgeons during teaching periods to assess the skills of the trainees.
In fact, a recent study~\cite{mota2018video} showed that residents spend more time viewing videos than specialists, highlighting the need for young surgeons to fully benefit from the procedure.
In~\cite{herrera2016the}, the authors showed that knot-tying scores and times for task completion improved significantly for the subjects that watched the videos of their own performance. 

However, when the trainees are willing to asses their progress over several trials of the same surgical task by re-watching their recorded surgical videos simultaneously, the problem of videos being out-of-synch makes the comparison between different trials very difficult if not impossible.  
This problem is encountered in many real life case studies, since experts on average complete the surgical tasks in less time than novice surgeons~\cite{mcNatt2001}. 
Thus, when trainees do enhance their skills, providing them with a feedback that pinpoints the reason behind the surgical skill improvement becomes problematic since the recorded videos exhibit different duration and are not perfectly aligned.  

Although synchronizing videos has been the center of interest for several computer vision research venues, contributions are generally focused on a special case where multiple simultaneously recorded videos (with different characteristics such as viewing angles and zoom factors) are being processed~\cite{wolf2002sequence,wedge2005trajectory,padua2010linear}.
Another type of multiple video synchronization uses hand-engineered features (such as points of interest trajectories) from the videos~\cite{wang2014videosnapping,evangelidis2011efficient}, making the approach highly sensitive to the quality of the extracted features.  
This type of techniques was highly effective since the raw videos were the only source of information available, whereas in our case, the use of robotic surgical systems enables capturing an additional type of data: the kinematic variables such as the $x,y,z$ Cartesian coordinates of the Da Vinci's end effectors~\cite{gao2014jhu}. 

In this paper, we propose to leverage the sequential aspect of the recorded kinematic data from the Da Vinci surgical system, in order to synchronize their corresponding video frames by aligning the time series data (see Figure~\ref{fig:vid-synch} for an example). 
When aligning two time series, the off-the-shelf algorithm is Dynamic Time Warping (DTW)~\cite{sakoe1978dynamic} which we indeed used to align two videos. 
However, when aligning multiple sequences, the latter technique does not generalize in a straightforward and computationally feasible manner~\cite{petitjean2014dynamic}. 
Hence, for multiple video synchronization, we propose to align their corresponding time series to the average time series, computed using the DTW Barycenter Averaging (DBA) algorithm~\cite{petitjean2014dynamic}.
This process is called Non-Linear Temporal Scaling (NLTS) and has been proposed to find the multiple alignment of a set of discretized surgical gestures~\cite{forestier2014non}, which we extend in this work to continuous numerical kinematic data.
Figure~\ref{fig:trials} depicts an example of stretching three different time series using the NLTS algorithm.   
Examples of the synchronized videos and the associated code can be found on our GitHub repository\footnote{\url{https://github.com/hfawaz/aime19}}, where we used the JHU-ISI Gesture and Skill Assessment Working Set (JIGSAWS)~\cite{gao2014jhu} to validate our work. 

The rest of the paper is organized as follows: in Section~\ref{sec-method}, we explain in details the algorithms we have used in order to synchronize the kinematic data and eventually their corresponding video frames. 
In Section~\ref{sec-ex}, we present our experiments and finally conclude the paper and discuss our future work in Section~\ref{sec-conc}.






\section{Methods} \label{sec-method}

In this section, we detail each step of our video synchronization approach. 
We start by describing the Dynamic Time Warping (DTW) algorithm which allows us to align two videos. 
Then, we describe how Non-Linear Temporal Scaling (NLTS) enables us to perform multiple video synchronization with respect to the reference average time series computed using the DTW Barycenter Averaging (DBA) algorithm. 

\subsection{Dynamic Time Warping}
Dynamic Time Warping (DTW) was first proposed for speech recognition when aligning two audio signals~\cite{sakoe1978dynamic}. 
Suppose we want to compute the dissimilarity between two time series, for example two different trials of the same surgical task, $A=(a_1,a_2,\dots,a_m)$ and $B=(b_1,b_2,\dots,b_n)$.
The length of $A$ and $B$ are denoted respectively by $m$ and $n$, which in our case correspond to the surgical trial's duration.  
Here, $a_i$ is a vector that contains six real values, therefore $A$ and $B$ can be seen as two distinct Multivariate Time Series (MTS). 

To compute the DTW dissimilarity between two MTS, several approaches were proposed by the time series data mining community~\cite{shokoohi2017generalizing}, however in order to apply the subsequent algorithm NLTS, we adopted the ``dependent'' variant of DTW where the Euclidean distance is used to compute the difference between two instants $i$ and $j$. 
Let $M(A,B)$ be the $m\times n$ point-wise dissimilarity matrix between $A$ and $B$, where $M_{i,j}=||a_i-b_j||^2$. 
A warping path $P=((c_1,d_1),(c_2,d_2),\dots,(c_s,d_s))$ is a series of points that define a crossing of $M$. 
The warping path must satisfy three conditions: (1) $(c_1,d_1)=(1,1)$; (2) $(c_s,d_s)=(m,n)$; (3) $0\le c_{i+1}-c_i \le 1$ and $0\le d_{j+1}-d_j \le 1$ for all $i<m$ and $j<n$. 
The DTW measure between two series corresponds to the path through $M$ that minimizes the total distance. 
In fact, the distance for any path $P$ is equal to $D_P(A,B)=\sum_{i=1}^{s}P_i$.
Hence if $\textbf{P}$ is the space of all possible paths, the optimal one - whose cost is equal to $DTW(A,B)$ - is denoted by $P^*$ and can be computed using: $\min_{P\in \textbf{P}}D_P(A,B)$. 

The optimal warping path can be obtained efficiently by applying a dynamic programming technique to fill the cost matrix $M$. 
Once we find this optimal warping path between $A$ and $B$, we can deduce how each time series element in $A$ is linked to the elements in $B$. 
We propose to exploit this link in order to identify which time stamp should be duplicated in order to align both time series, and by duplicating a time stamp, we are also duplicating its corresponding video frame. 
Concretely, if elements $a_i$, $a_{i+1}$ and $a_{i+2}$ are aligned with the element $b_j$ when computing $P^*$, then by duplicating twice the video frame in $B$ for the time stamp $j$, we are dilating the video of $B$ to have a length that is equal to $A$'s. 
Thus, re-aligning the video frames based on the aligned Cartesian coordinates: if subject $S_1$ completed ``\textit{inserting the needle}'' gesture in 5 seconds, whereas subject $S_2$ performed the same gesture within 10 seconds, our algorithm finds the optimal warping path and duplicates the frames for subject $S_1$ in order to synchronize with subject $S_2$ the corresponding gesture.     
Figure~\ref{fig:vid-synch} illustrates how the alignment computed by DTW for two time series can be used in order to duplicate the corresponding frames and eventually synchronize the two videos.  

\subsection{Non-Linear Temporal Scaling}

The previous DTW based algorithm works perfectly when synchronizing only two surgical videos. 
The problem arises when aligning three or more surgical trials simultaneously, which requires a multiple series alignment. 
The latter problem has been shown to be NP-Complete~\cite{wang1994} with the exact solution requiring $O(L^N)$ operations for $N$ sequences of length $L$. 
This is clearly not feasible in our case where $L$ varies between $10^3$ and $10^4$ and $N\ge3$, which is why we ought to leverage an approximation of the multiple sequence alignment solution provided by the DTW Barycenter Averaging (DBA) algorithm which we detail in the following paragraph. 

DBA was originally proposed in~\cite{petitjean2011a} as a technique that averages a set of time series by leveraging an approximated multiple sequence alignment algorithm called Compact Multiple Alignment (CMA)~\cite{petitjean2012summarizing}.
DBA iteratively refines an average time series $T$ and follows an expectation-maximization scheme by first considering $T$ to be fixed and finding the best CMA between the set of sequences $D$ (to be averaged) and the refined average sequence $T$. 
After computing the CMA, the alignment is now fixed and the average sequence $T$ is updated in a way that minimizes the sum of DTW distances between $T$ and $D$~\cite{petitjean2014dynamic}.  

DBA requires an initial value for $T$. 
There exist many possible initializations for the average sequence~\cite{petitjean2012summarizing}, however, since our ultimate goal is to synchronize a set of sequences $D$ by duplicating their elements (dilating the sequences), we initialize the average $T$ to be equal to the longest instance in $D$.
We then find precisely the exact optimal number of time series elements - and their associated video frames - to be duplicated in order to synchronize multiple videos, using the NLTS technique which we describe in details in the following paragraph. 

Non-Linear Temporal Scaling (NLTS) was originally proposed for aligning discrete sequences of surgical gestures~\cite{forestier2014non}.
In this paper, we extend the technique for numerical continuous sequences (time series). 
The goal of this final step is to compute the approximated multiple alignment of a set of sequences $D$ which will eventually contain the precise information on how much a certain frame from a certain series should be duplicated. 
We first start by computing the average sequence $T$ (using DBA) for a set of time series $D$ that we want to align simultanously. 
Then, by recomputing the Compact Multiple Alignment (CMA) between the refined average $T$ and the set of time series $D$, we can extract an alignment between $T$ and each sequence in $D$. 
Thus, for each time series in $D$ we will have the necessary information (extracted from CMA) in order to dilate the time series appropriately to have a length that is equal to $T$'s, which also corresponds to the length of the longest time series in $D$.   
Figure~\ref{fig:trials}, depicts an example of aligning three different time series using the NLTS algorithm. 

\section{Experiments}\label{sec-ex}
We start by describing the JIGSAWS dataset we have used for evaluation, before presenting our experimental study. 

\subsection{Dataset}

\begin{figure*}[t]
  \centering
  \includegraphics[width=.325\linewidth]{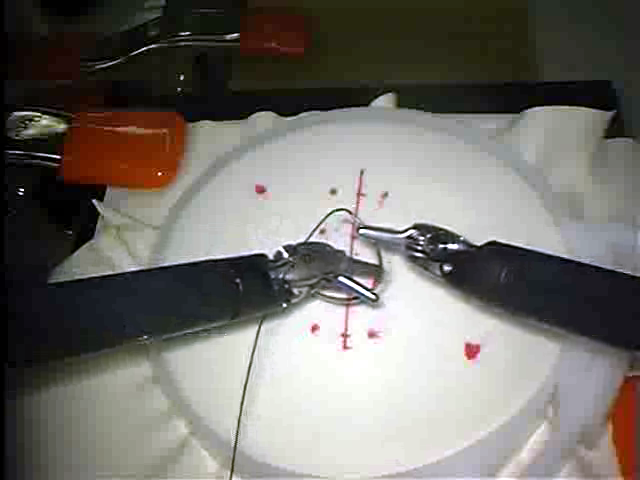}
   \includegraphics[width=.325\linewidth]{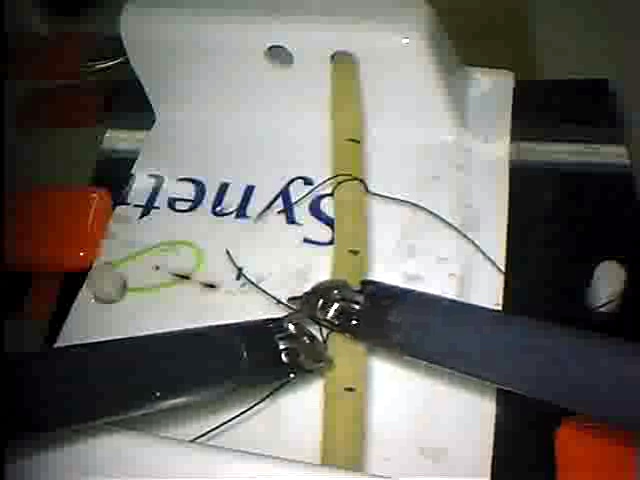}
   \includegraphics[width=.325\linewidth]{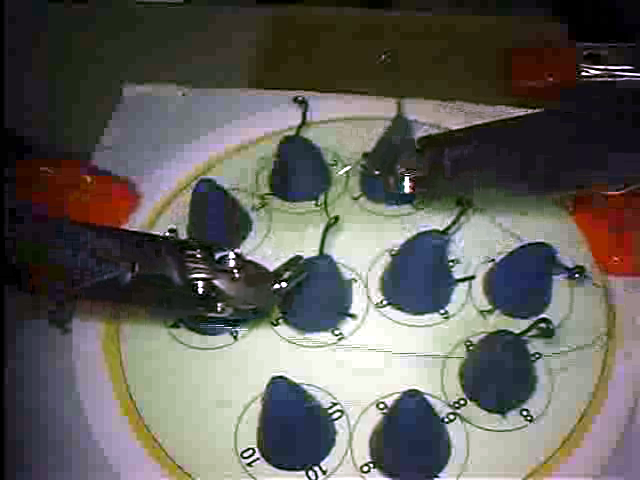}
  \caption{Snapshots of the three surgical tasks in the JIGSAWS dataset (from left to right): suturing, knot-tying, needle-passing \cite{gao2014jhu}.}
  \label{fig:jig}
\end{figure*}

The JIGSAWS dataset~\cite{gao2014jhu} includes data for three basic surgical tasks performed by study subjects (surgeons). 
The three tasks (or their variants) are usually part of the surgical skills training program. 
Figure~\ref{fig:jig} shows a snapshot example for each one of the three surgical tasks (Suturing, Knot Tying and Needle Passing). 
The JIGSAWS dataset contains kinematic and video data from eight different subjects with varying surgical experience: two experts (E), two intermediates (I) and four novices (N) with each group having reported respectively more than 100 hours, between 10 and 100 hours and less than 10 hours of training on the Da Vinci. 
All subjects were reportedly right-handed. 

The subjects repeated each surgical task five times and for each trial the kinematic and video data were recorded. 
When performing the alignment, we used the kinematic data which are numeric variables of four manipulators: left and right masters (controlled directly by the subject)  and  left  and  right slaves (controlled indirectly by the subject via the master manipulators).
These kinematic variables (76 in total) are captured at a frequency equal to 30 frames per second for each trial.
Out of these 76 variables, we only consider the Cartesian coordinates ($x,y,z$) of the left and right slave manipulators, thus each trial will consist of an MTS with 6 temporal variables. 
We chose to work only with this subset of kinematic variables to make the alignment coherent with what is visible in the recorded scene: the robots' end-effectors which can be seen in Figure~\ref{fig:jig}. 
However other choices of kinematic variables are applicable, which we leave the exploration for our future work.
Finally we should mention that in addition to the three self-proclaimed skill levels (N,I,E) JIGSAWS contains the modified Objective Structured Assessment of Technical Skill (OSATS) score~\cite{gao2014jhu}, which corresponds to an expert surgeon observing the surgical trial and annotating the performance of the trainee. 

\subsection{Results}

\begin{figure}[t]
    \centering
    \subfloat[Videos synchronization process]{
    \includegraphics[width=.58\linewidth]{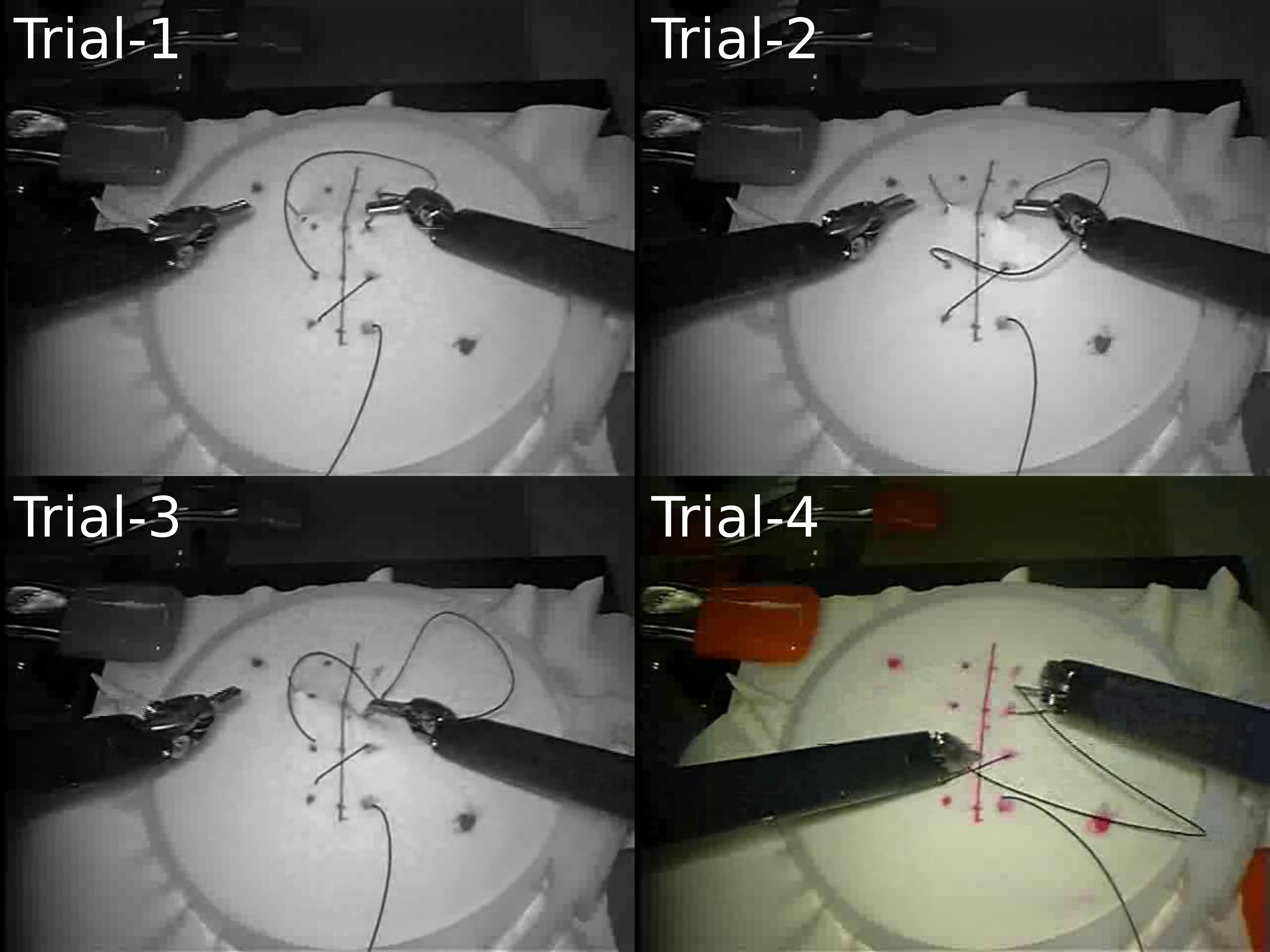}
    \label{sub:video-4x4-unsynched}
    }\\
    \subfloat[Perfectly aligned videos]{
 \includegraphics[width=.58\linewidth]{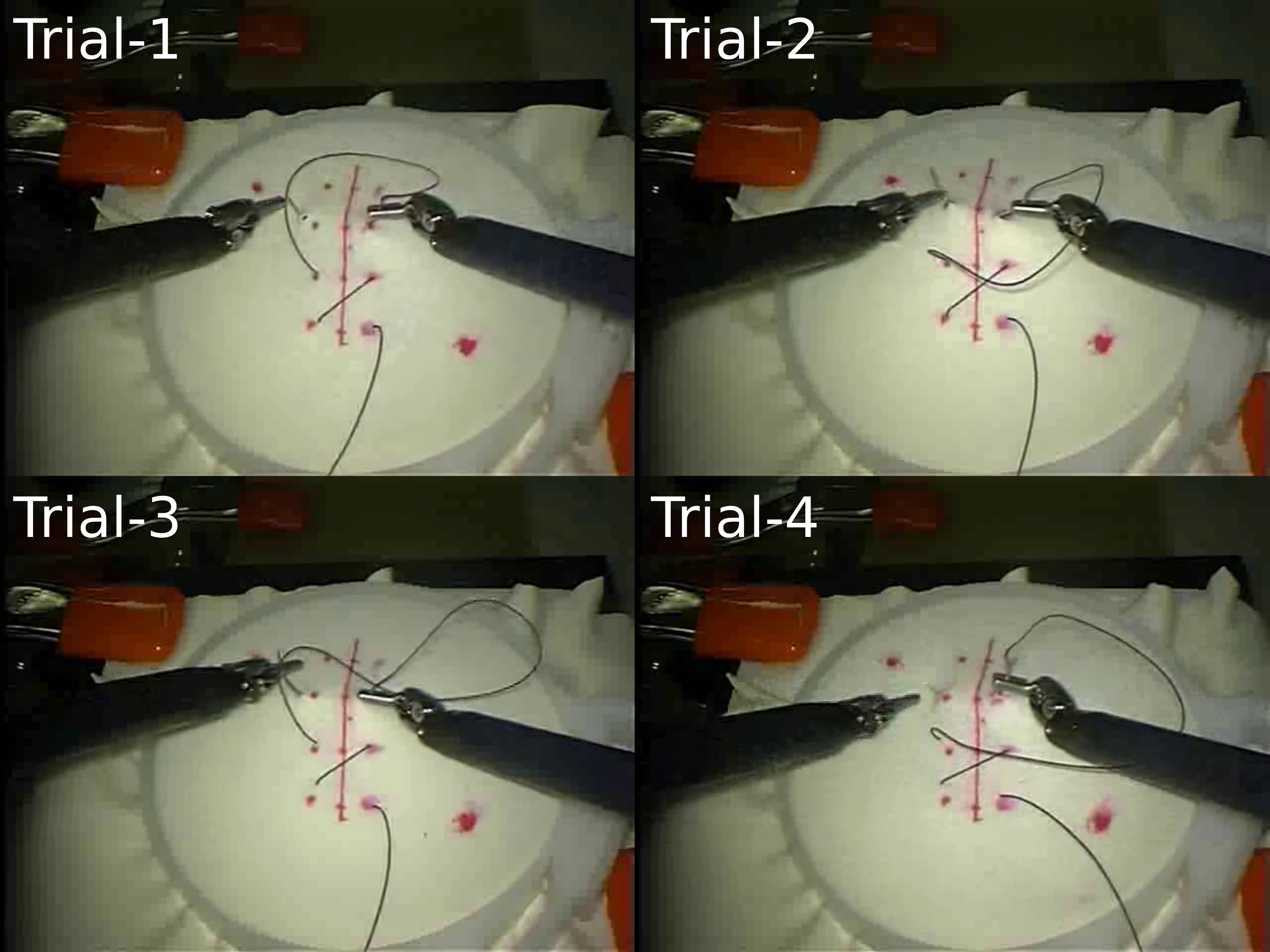}
 \label{sub:video-4x4-synched}
      }
    \caption{Video alignment procedure with duplicated (gray-scale) frames.}
    \label{fig:video-4x4}
\end{figure}

We have created a companion web page\footnote{\url{https://germain-forestier.info/src/aime2019/}} to our paper where several examples of synchronized videos can be found. 
Figure~\ref{fig:video-4x4} illustrates the multiple videos alignment procedure using our NLTS algorithm, where gray-scale images indicate duplicated frames (paused video) and colored images indicate a surgical motion (unpaused video).  
In Figure~\ref{sub:video-4x4-unsynched} we can clearly see how the gray-scale surgical trials are perfectly aligned. 
Indeed, the frozen videos show the surgeon ready to perform ``\textit{pulling the needle}'' gesture~\cite{gao2014jhu}.
On the other hand, the colored trial (bottom right of Figure~\ref{sub:video-4x4-unsynched}) shows a video that is being played, where the surgeon is performing ``\textit{inserting the needle}'' gesture in order to catch up with the other paused trials in gray-scale.
Finally, the result of aligning simultaneously these four surgical trials is depicted in Figure~\ref{sub:video-4x4-synched}. 
By observing the four trials, one can clearly see that the surgeon is now performing the same surgical gesture ``\textit{pulling the needle}'' simultanously for the four trials.
We believe that this type of observation will enable a novice surgeon to locate which surgical gestures still need some improvement in order to eventually become an expert surgeon. 

\begin{figure}
    \centering
    \includegraphics[width=0.55\linewidth]{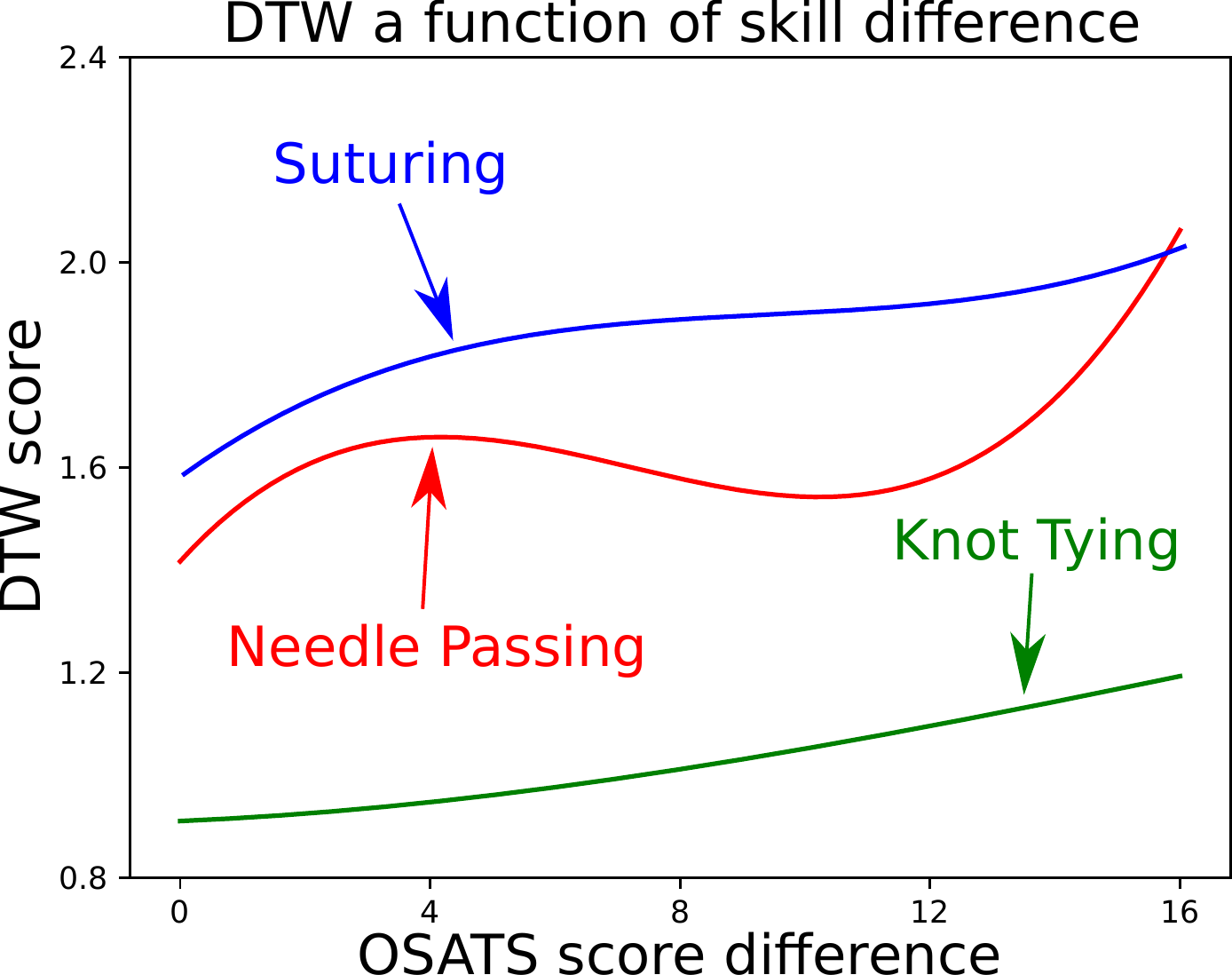}
    \caption{A polynomial fit (degree 3) of DTW dissimilarity score (y-axis) as a function of the OSATS score difference between two surgeons (x-axis).}
    \label{fig:dtw-osats}
\end{figure}

Furthermore, in order to validate our intuition that DTW is able to capture characteristics that are in relationship with the motor skill of a surgeon, we plotted the DTW distance as a function of the OSATS~\cite{gao2014jhu} score difference. 
For example, if two surgeons have both an OSATS score of 10 and 16 respectively, the corresponding difference is equal to $|10-16|=6$. 
In Figure~\ref{fig:dtw-osats}, we can clearly see how the DTW score increases whenever the OSATS score difference increases. 
This observation suggests that the DTW score is low when both surgeons exhibit similar dexterity, and high whenever the trainees show different skill levels. 
Therefore, we conclude that the DTW score can serve as a heuristic for estimating the quality of the alignment (whenever annotated skill level is not available) - especially since we observed low quality alignments for surgeons with very distinct surgical skill levels.

Finally, we should note that this work is suitable for many research fields involving motion kinematic data with their corresponding video frames. 
Examples of such medical applications are assessing mental health from videos~\cite{yamada2017detecting} where wearable sensor data can be seen as time series kinematic variables and leveraged in order to synchronize a patient's videos and compare how well the patient is responding to a certain treatment. 
Following the same line of thinking, this idea can be further applied to kinematic data from wearable sensors coupled with the corresponding video frames when evaluating the Parkinson's disease evolution~\cite{criss2011} as well as infant grasp skills~\cite{li2019manipulation}.

\section{Conclusion}\label{sec-conc}
In this paper, we showed how kinematic time series data recorded from the Da Vinci's end effectors can be leveraged in order to synchronize the trainee's videos performing a surgical task. 
With personalized feedback during surgical training becoming a necessity~\cite{IsmailFawaz2018e,forestier2018surgical}, we believe that replaying \emph{synchronized} and well \emph{aligned} videos would benefit the trainees in understanding which surgical gestures did or did not improve after hours of training, thus enabling them to further reach higher skills and eventually become experts. 
We acknowledge that this work needs an experimental study to quantify how beneficial is replaying synchronized videos for the trainees versus observing non-synchronized trials.
Therefore, we leave such exploration and clinical try outs to our future work.  

\bibliographystyle{splncs04}
\bibliography{biblio}

\end{document}